\documentclass[10pt,twocolumn,letterpaper]{article}

\usepackage[pagenumbers]{cvpr} 

\usepackage{graphicx}
\usepackage{amsmath}
\usepackage{amssymb}
\usepackage{booktabs}
\usepackage{enumitem}

\usepackage{algorithm}
\usepackage{algorithmicx}
\usepackage{algpseudocode}

\usepackage[table,xcdraw]{xcolor}
\usepackage{multirow}

\usepackage[pagebackref,breaklinks,colorlinks]{hyperref}

\usepackage[capitalize]{cleveref}
\crefname{section}{Sec.}{Secs.}
\Crefname{section}{Section}{Sections}
\Crefname{table}{Table}{Tables}
\crefname{table}{Tab.}{Tabs.}

\def\cvprPaperID{3227} 
\def\confName{CVPR}
\def\confYear{2023}

\begin{document}

\title{Adversarial Attack with Raindrops}

\author{Jiyuan Liu, Bingyi Lu, Mingkang Xiong, Tao Zhang, Huilin Xiong\\
Shanghai Jiao Tong University\\
Shanghai, China\\
}
\maketitle

\begin{abstract}
Deep neural networks (DNNs) are known to be vulnerable to adversarial examples, which are usually designed artificially to fool DNNs, but rarely exist in real-world scenarios. In this paper, we study the adversarial examples caused by raindrops, to demonstrate that there exist plenty of natural phenomena being able to work as adversarial attackers to DNNs. Moreover, we present a new approach to generate adversarial raindrops, denoted as AdvRD, using the generative adversarial network (GAN) technique to simulate natural raindrops. The images crafted by our AdvRD look very similar to the real-world raindrop images, statistically close to the distribution of true raindrop images, and more importantly, can perform strong adversarial attack to the state-of-the-art DNN models.  On the other side, we show that the adversarial training using our AdvRD images can significantly improve the robustness of DNNs to the real-world raindrop attacks. Extensive experiments are carried out to demonstrate that the images crafted by AdvRD are visually and statistically close to the natural raindrop images, can work as strong attackers to DNN models, and also help improve the robustness of DNNs to raindrop attacks. 

\end{abstract}

\section{Introduction}
\label{sec:intro}

Deep neural networks have achieved outstanding performance in many computer vision tasks, such as image classification~\cite{Resnet,InceptionV3}, object detection~\cite{YOLOv2,RCNN}, and semantic segmentation~\cite{FCN}. However, they are also vulnerable to various adversarial attacks, which are specifically designed to mislead DNNs with some invisible perturbations. The perturbed images, called adversarial examples, have attracted a lot of research interests and attention, since they could potentially threaten many safety-critical applications, including medical diagnosis~\cite{medical-diagnosis} and autonomous driving~\cite{autonomous-cars}. 
Existing adversarial attacks can be divided into two categories: 1) digital attacks, where the adversarial perturbations are crafted in the digital domain, \eg, the conventional gradient-based adversarial attacks. 2) physical attacks, where the perturbations are crafted on real existing objects, such as road sign, T-shirt, \etc, to achieve the attacking goal. 

Since digitally crafted adversarial examples rarely exist in the real-world environment, physical attacks have recently drawn more and more attention. One popular strategy adopted by physical attacks is to add some carefully designed artifacts on the target object, \eg, stickers on Stop signs~\cite{physical_stopsign} or colorful patterns on eyeglass frames~\cite{physical_glasses}. However, these attacks often generate unnatural textures, which are quite visible to human eyes. Thus many works focus on generating adversarial examples with natural styles that appear legitimate to human eyes, \eg, adversarial shadows~\cite{physical_shadows} caused by polygons. Nevertheless, these visually valid adversarial examples are still artifacts and seldom appear in real-world environments.

In recent years, some researchers~\cite{advrain,advhaze} propose to employ natural phenomena, such as rain and haze, to generate more natural adversarial attacks. For example, Zhai \etal~\cite{advrain} generates adversarial rain with different angles and intensities to cheat the target DNN. Gao \etal~\cite{advhaze} proposes adversarial haze to attack the DNNs. Although the adversarial weather examples crafted in these works show strong attack ability to DNNs, they are not real rain or haze, and often look unnatural, either because the models used to simulate weather phenomena are not sophisticated, or because too much attention is put on the attack strength of the adversarial examples, and their reality is ignored. 

In this paper, we investigate the adversarial examples caused by raindrops to show a fact that there exist a lot of natural phenomena, like raindrops, being able to work as adversarial attackers to DNNs. Therefore, it is crucial for such type of applications as autonomous driving to find an effective way to defend against these natural adversarial attacks. For this purpose, we propose a scheme, denoted as AdvAD, to generate adversarial raindrop images based on the GAN technique. Specifically, AdvAD trains a Generative Adversarial Network (GAN) on a real-world raindrop dataset~\cite{raidrop_removal_cvpr2018} until it can transform a clean input image into a natural raindrop style image, and meanwhile, a transfer learning classifier is embedded in the GAN framework to endue the generated raindrop images more power of adversarial attacking. The adversarial raindrop images generated by our scheme are shown to be very similar to the natural raindrop images, not only from human viewpoint, but also from the statistic measure of two distributions. Finally, we show that the adversarial raindrop images help improve the robustness of DNN models to the attacks of natural adversarial raindrops. Extensive experiments are carried out to demonstrate the effectiveness of our scheme. 
 
The main contributions of our work are the following:
\begin{itemize}	
	\item We propose a novel approach, based on the GAN technique, to generate adversarial raindrop images that are visually and statistically similar to the natural raindrop images. 
	\item We show real-world raindrops can act as adversarial examples to mislead DNNs, which could bring substantive threats to such security-critical applications as autonomous driving. Adversarial training using our AdvRD samples can help improve the robustness of DNNs to real-world raindrop perturbations. 
	
\end{itemize}

\section{Related Work}
\label{sec:Rela}  

\subsection{Adversarial Attacks}

The adversarial perturbation phenomenon was first found by Szegedy \etal~\cite{L-BFGS}, in which carefully manipulated perturbations were added to the original images to fool DNN classifiers. The perturbed images, named adversarial examples, are usually imperceptible to humans, but could mislead the classifier to output incorrect predictions. Let $x\in \mathbb{R} ^d$ denote a clear image sample and $x'$ its corresponding adversarial example. Adversarial perturbation can be expressed as an optimizing problem:
\begin{equation}
	\underset{x'}{arg\max}\,\,L_C\left( x',y \right) \,\,s.t. \left\| x'-x \right\|_{p} <\epsilon, 
	\label{eq:label}
\end{equation}
where $y$ is the ground-truth label of $x$ and $\epsilon$ is the perturbation budget. $L_C$ denotes the loss function of the classifier. The perturbation $\delta = x'-x$ is often bounded by $\ell_{p}$-norms to guarantee their imperceptibility to human eyes. $\ell_{2}$, $\ell_{\infty}$ and $\ell_{0}$ are commonly used norms. In the literature, plenty of adversarial attacks have been proposed, which can be divided into two categories, digital attacks and physical attacks, according to the domain that the adversarial perturbations are crafted in.

\subsection{Digital Attacks}

Adversarial attack in the digital domain is to craft perturbation for each pixel of the input image to mislead DNNs' prediction. According to the transparency of the target DNN model to the attacker, digital attacks can be grouped into white-box~\cite{FGSM,BIM,CW} attacks and black-box attacks~\cite{MI-FGSM,DIM,TIM,NIM,NATTACK,Zoo}. In white-box attacks, attackers have access to entire information of the target model, and therefore, can design efficient ways of using the model gradients, like those in PGD~\cite{PGD}, to perturb image to achieve their goal. In black-box scenarios, only outputs of the target model are accessible. Hence attackers need to estimate the gradient by querying the target model~\cite{NES,SPSA} or rely on the adversarial transferability~\cite{DIM,TIM}. Although the adversarial examples crafted in the digital domain work well in cheating DNN models, they in fact rarely exist in the real-world.  

\subsection{Physical Attacks}
Kurakin \etal~\cite{Physical_first_attack} first demonstrates that adversarial examples also exist in real-world environments. They find that hard copies of the digital adversarial examples can still fool the target model. One popular strategy of physical attacks is ``sticker-pasting", which attaches adversarial patches to the object to mislead DNN models. For example, Eykholt \etal~\cite{physical_stopsign_cvpr2018} put adversarial patches on the road sign to fool a well-trained auto vehicle. Xu \etal~\cite{physical_T_shirt} print the adversarial patch on T-shirt to help a person escape from detection model. Sharif \etal~\cite{physical_glasses} apply the adversarial textures on eye-glasses frames to fool facial recognition system. 

Another line of work tries to attack the target model in a non-invasive manner. Applying semitransparent stickers~\cite{physical_sticker_recognize_ICML_2019,physical_translucent_patch_cvpr2021} to a camera lens is a simple yet effective strategy to fool recognition and detection systems. Some methods utilize optic devices to generate adversarial examples. Gnanasambandam \etal~\cite{physical_optical_projector_iccv2021} use a projector to perform the adversarial perturbations on the object to realize the attack. Sayles \etal~\cite{physical_optical_light_cvpr2021} illuminate the object with a modulated light signal to craft adversarial examples invisible to humans. Duan \etal~\cite{physical_optical_laser_cvpr2021} shoot a laser beam onto the target object to achieve a fast attack.

Recently, some works try to camouflage physical-world adversarial examples in natural phenomena. Zhai \etal~\cite{advrain} transform clean images into rainy styles to perform attacks. Gao \etal~\cite{advhaze} camouflage perturbations into the haze to mislead the classifier. Zhong \etal~\cite{physical_optical_shadow_cvpr2022} utilize perturbations visually like shadows to craft adversarial examples. These works generate visually natural images based on theoretical models. However, the synthetic samples are not truly compatible aligned with their counterparts in the real world.

\begin{figure*}[!t]
	\centering
	\includegraphics[width=0.9\linewidth]{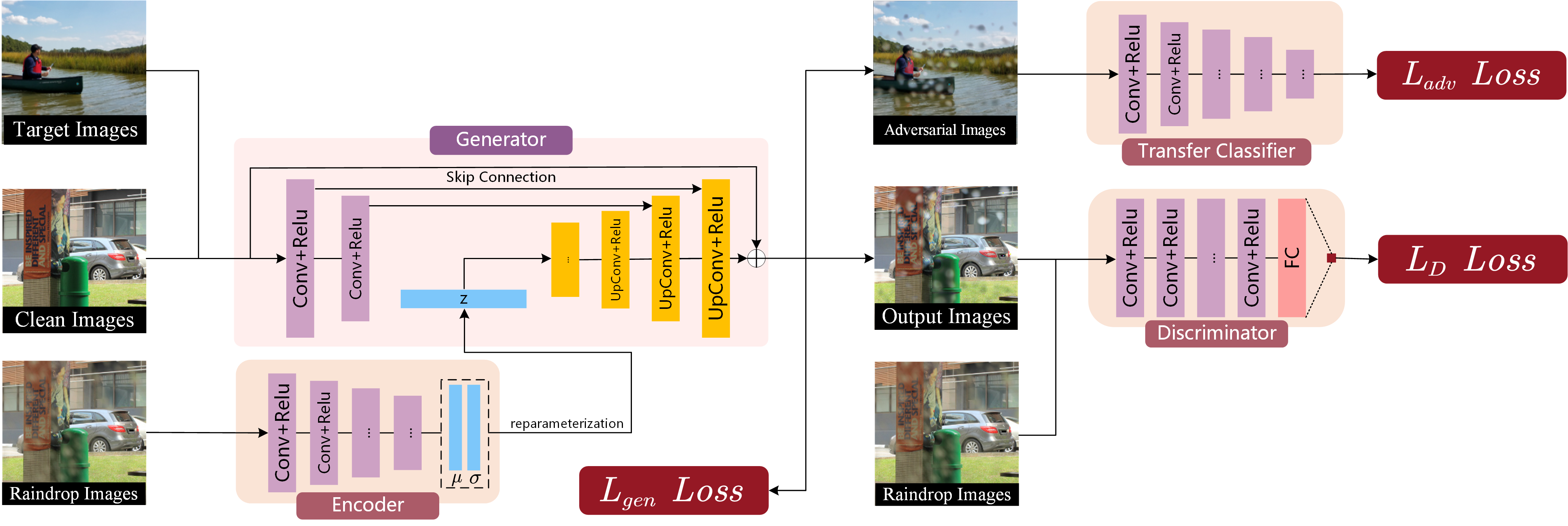}
	\caption{Illustrations of the pipline of raindrop generation network.}
	\label{fig:total}
\end{figure*} 

\section{Methodology}     
In this section, we present our approach to generate adversarial raindrops in both digital and physical domains. We first describe a physical way to acquire real-world adversarial raindrop images, and show their strong capability to mislead well-trained DNN models. Then, we focus our attention on how to generate adversarial raindrops in the digital domain, using a quasi-GAN technique.   

\subsection{Real-World Adversarial Raindrops}
\label{real-world raindrop}
When we are taking pictures on a rainy day, sometimes through a glass of window, there are always a few of raindrops attached to our camera lens or the window glass. What we finally acquire are raindrop images, which could mislead well-trained DNN models. To investigate how often the real-world raindrop images mislead a pre-trained DNN model, we follow the strategy of \cite{raidrop_removal_cvpr2018} to acquire raindrop images, and statistically estimate the possibility that a raindrop image misleads a well-trained DNN classifier.

\begin{figure}[!t]
	\centering
	
	\includegraphics[width=\linewidth]{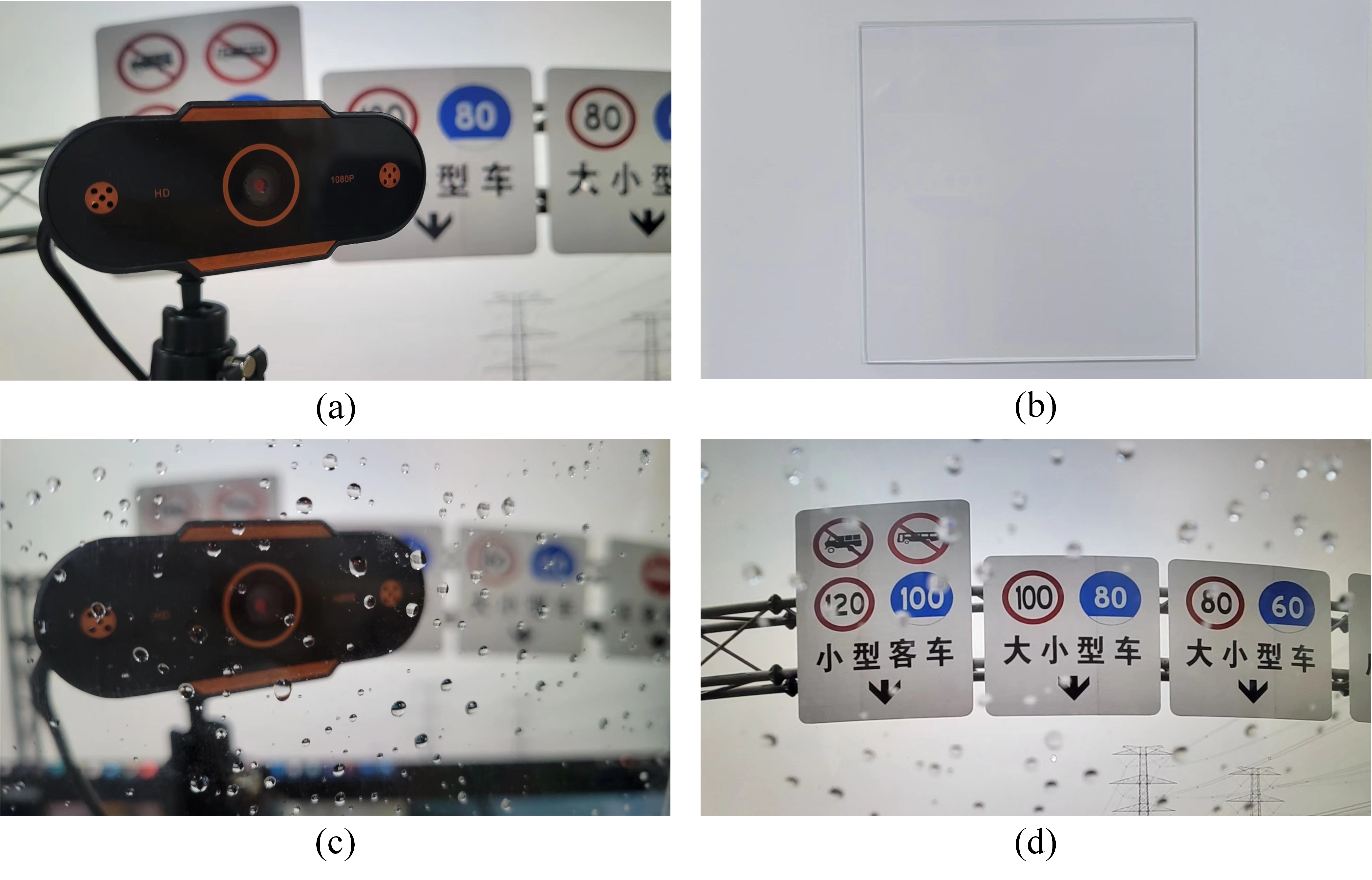}
		\setlength{\abovecaptionskip}{0.0cm}
		\setlength{\belowcaptionskip}{-0.4cm}

	\caption{(a) the camera used to capture images; (b) the glass used to craft raindrops; (c) raindrops randomly sprayed on the glass in front of camera lens; and (d) final raindrop image.}
	\label{fig:lab}
\end{figure}

We randomly spray some small drops of water on a glass and put it in front of the camera's lens. The clean images are shown on the computer screen one by one. We fix the position of the camera and the screen, and randomly move and rotate the glass to collect 5 seconds of video for each image. If we find at least one frame in the video that misleads a pre-trained DNN model, we call this frame a real-world adversarial raindrop image of the DNN model. Fig.~\ref{fig:lab} illustrates the process of obtaining real-world adversarial raindrops. We find that there is a good chance (over 50\%, see Sec.~\ref{physical-attack}) to get a real-world adversarial raindrop image for a given DNN model. Obviously, it is also dependent on the size and density of the water drops. More detailed information and experimental results are presented in Sec.~\ref{physical-attack}.

\subsection{Adversarial Raindrops in Digital Domain}
Adversarial training is considered as the most effective way to improve the robustness of DNN models to adversarial attacks. It also works for the adversarial raindrops. However, it needs to train the classifier with large number of adversarial raindrop images, which are hard to collect in practice. Therefore, we proposed a novel approach, AdvRD, to simulate natural raindrops in the digital domain using a quasi-GAN framework. The generated raindrops are shown to be considerably strong in attacking DNN models, and similar to the natural raindrops from viewpoints of human vision and statistical analysis (see Sec.~\ref{realness}).        

Fig.~\ref{fig:total} shows the illustrations of training stages for the proposed raindrop generation networks. we generate adversarial raindrop images based on a quasi-GAN framework, which contains three sub-networks, generator, discriminator, and a transfer classifier. Different from the conventional GAN, the generator $G$ in our GAN architecture tries to generate raindrop images as real as possible, aiming to cheat not only the discriminator, but also the transfer classifier. The discriminator $D$ tries to identify whether the input image is a real raindrop image or from the generator. The transfer classifier $C$ is employed to endue the generated raindrop images with adversarial attacking capability.

Our generative adversarial loss can be formulated as:   
\begin{equation}
	\begin{split}
		\underset{G}{\min}\underset{D}{\max}~V\left( G,D \right) &=\mathbb{E} _{o\sim P_{raindrop}}\left[ \log \left( D\left( o \right) \right) \right] \\
		&+\mathbb{E} _{b\sim P_{clean}}\left[ \log \left( 1-D\left( G\left( b,z \right) \right) \right) \right] 
		\\
		&+\mathbb{E} _{x\sim P_{pred}} \left[\left\| L_C\left( G\left( x,z \right) \right) -\eta \right\| _1 \right] 
	\end{split}
	\label{eq:whole-loss}
\end{equation}   
where $o$ is a sample drawn from the ground-truth raindrop images and $b$ is the corresponding image of clean natural images. $z$ is a Gaussian noise vector and $x$ is a clean image that is correctly classified by $C$. $\eta$ is a factor to balance the reality and attacking ability of the generated raindrops.

\subsubsection{Generative Network}
The first goal of the generator is to generate real-like raindrop images to fool the discriminator. To simulate natural raindrops, the generator should consider the background scene during the raindrop generation. As shown in Fig.~\ref{fig:total}, we apply several convolutional layers to extract the shallow feature of clean images and fuse them with the intermediate features calculated from a noise vector $z$, to obtain the final raindrops. This process can be expressed as follows:
\begin{equation}
	\begin{aligned}
		o' &= G\left( b ,z \right)\\
		&= G(b,E(o)) ,		
	\end{aligned}
	\label{eq:vae-generate}
\end{equation} 
where $o'$ represents the synthetic raindrop images, and $E$ represents the encoder. It converts the raindrop image into features, \ie, the mean and variance for the latent variable $z$.

We use pairs of images with and without raindrops, $\left\{ o_n,b_n \right\} _{n=1}^{N}$, to train the generative network. The generative loss $L_{G}$ for the first goal of generator is expressed as:  
\begin{equation} 
	L_G=L_{gen}+\alpha _{1}\cdot L_z+\alpha _{2}\cdot L_p,
	\label{eq:weather generation loss}
\end{equation}
where the first term of Eq.~(\ref{eq:weather generation loss}), $L_{gen}$, intends to train the generator to output raindrop images that can fool the discriminator. This loss is computed by:
\begin{equation} 
	L_{gen}=log(1-D(o')) .
	\label{eq:Lgen}
\end{equation}

The second term of Eq.~(\ref{eq:weather generation loss}), $L_z$, constrains the latent variable $z$ encoded by $E$, to obey the isotropic Gaussian distribution, which is calculated as:    
\begin{equation}
	\begin{aligned} 
	L_{z} &= D_{KL}\left[ p\left( z \right) ||\mathcal{N} \left( 0,I \right) \right] \\
	&=\sum_{i=1}^d{\left[ \frac{\mu ^2_i}{2}+\frac{1}{2}\left( \sigma _i -\log \sigma _i -1 \right) \right]}
	\label{eq:Lz}
	\end{aligned}
\end{equation}
where $\mu$ and $\sigma$ are the mean and variance of the latent variable $z$, and $d$ denotes the dimension of $z$.

The last loss in Eq.~(\ref{eq:weather generation loss}), $L_p$, is adopted to make the generated raindrops more realistic. Inspired by the observation that raindrops only affect partial pixels of the image, thus we choose the $L_{1}$ norm to encourage the generator crafts sparse perturbations:    
\begin{equation} 
	L_p=\left\| o'-b \right\| _1
	\label{eq:Lp}
\end{equation}
In practice, we find that $L_p$ can help to improve the quality of synthetic raindrop images and accelerate the convergence speed of GAN.

\subsubsection{Discriminative Network} 
 The discriminator $D$ tries to identify if an input image comes from real data distribution rather than $G$. Thus the loss of discriminator is defined as: 

\begin{equation}
    L_D=- log(D\left( o \right)) -log(1-D(o'))	
	\label{eq:discriminator loss}
\end{equation}
Some works~\cite{Globally_local_GAN} require discriminator to identify global and local images to improve the performance of GAN, which needs complex design of loss functions and more computing resources. To balance the performance and efficiency of the GAN, we only use a global discriminator whose structure is mainly based on AlexNet~\cite{Alexnet}, since we experimentally find that this sample network can satisfyingly meet our needs. 
 
\begin{figure}[!t]
	\centering
	\includegraphics[width=\linewidth]{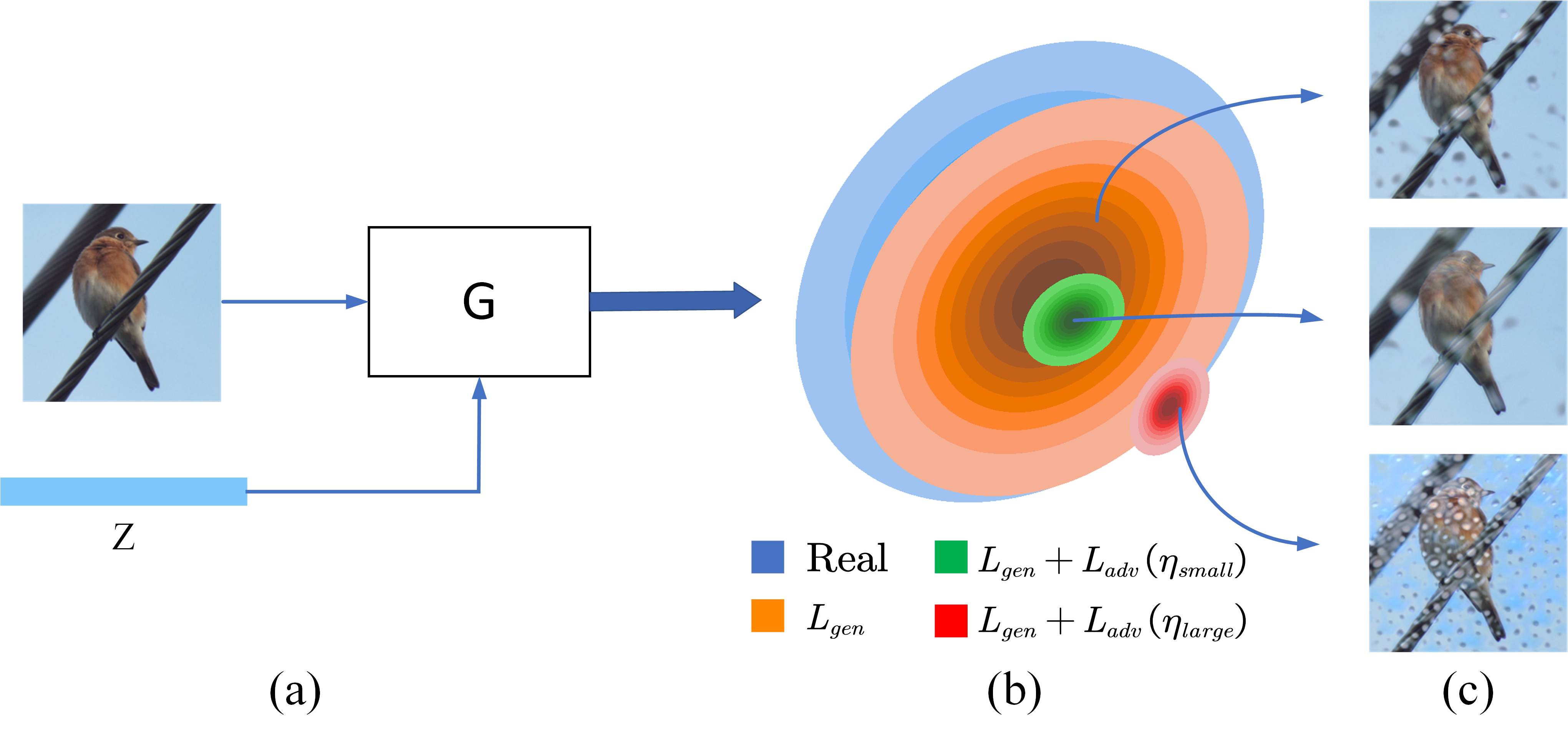}
	\setlength{\abovecaptionskip}{-0.1cm}
	\setlength{\belowcaptionskip}{-0.5cm}
	\caption{(a) The generator and the input. (b) The distributions learned by generators that trained by different losses. (c) Raindrops images sampled from different distributions. Raindrop images in the distribution trained with larger $\eta$ will fool the target classifier more easily but deviate from the real raindrop distribution.}
	\label{fig:Ladv effect distribution}
\end{figure}
\subsubsection{Transfer Classifier}
The second goal of the generator is to endue the generated raindrop images with more power of adversarial attack, since, according to our observation, only a small amount of raindrop images crafted from the previous generator G can successfully mislead a target DNN model. So, in our GAN architecture, a transfer learning network $C$, called transfer classifier, is added to the conventional GAN framework, aiming to transform the generated raindrop images to adversarial examples. Below is the loss function we used in the training:
\begin{equation}
	L_{adv}=\,\,\left\| L_C\left( G(x,z) ,y \right) -\eta \right\| _1, 
	\label{eq:untarget Lben}
\end{equation}
where $x$ is a clean image correctly classified by $C$, and $\eta$ is a positive constant used to limit the range of classification loss $L_C$, so that we can make a good trade-off between attacking ability and authenticity of the generated raindrops. Fig~\ref{fig:Ladv effect distribution} shows the effect of $\eta$ on the attacking ability and authenticity. More detailed analyses of $\eta$ are presented in Sec.~\ref{ablution}. 

Overall, the generative loss $L_{G}$ used during the training of GAN is expressed as:  
\begin{equation} 
	L_G=L_{gen}+\alpha _{1}\cdot L_z+\alpha _{2}\cdot L_p +\alpha _{3}\cdot L_{adv}.
	\label{eq:weather generation loss_stage2}
\end{equation}
\subsubsection{Adversarial Raindrop Attack}
After training of the GAN architecture, we fix its parameters, and generate adversarial raindrop images by solving the following optimization problem:
\begin{equation}
		\underset{z}{arg\max}\,\,L_{TC}\left( G\left( x,z \right) ,y \right), s.t. \left\| z \right\| <\epsilon_{z},
	\label{eq:Lc}
\end{equation}
where $L_{TC}$ is the loss function of the target DNN classifier, and $\epsilon_{z}$ denotes the threshold for $z$. In white-box scenario, Eq.~(\ref{eq:Lc}) can be solved by the gradient-descent method. The iteration formula can be written as follows:
\begin{equation}
	z_{t+1}=z_{{t}}+\alpha _{{z}}\cdot sign\left( \nabla _{z_{t}} L_{TC}\left( G\left( x,z \right) ,y \right) \right),  
	\label{eq:AWA_white}
\end{equation}
where $\alpha_{z}$ is the step size of the iteration. In black-box scenario, the gradients of loss \wrt the input noise are unavailable. Hence we adopt a simple but effective way to estimate the optimal $z$, that is, sampling $N$ inputs from the Gauss noise distribution and choose the one that can fool the target model. This strategy is similar to that in the query-based black-box attacking~\cite{Zoo,NATTACK}, and avoids the tough and complicated gradient estimation of the target model via a large number of queries. Experimentally, our method can achieve a fairly high black-box ASR (Attack Success Rate, 51.2\% for ResNet50) with a small query count $N=5$. The algorithm is denoted as AdvRD, and summarized in Algorithm~\ref{alg:AWA} for the case of white-box scenario. Note that the proposed method can generally combine with any gradient-based attacks to enhance its attack ability, \eg, MIM~\cite{MI-FGSM}, DIM~\cite{DIM}, and TIM~\cite{TIM}.
\begin{algorithm}
	\caption{Adversarial Raindrop Attack. (White-box)}\label{alg:AWA}
	\textbf{Input:} The target classifier $f$, loss function $L_{TC}$\\
	\qquad A original sample $x$, the ground-truth label $y$.\\
	 The raindrop generator $G$.\\
	 Noise sampling number $N$, iteration number $T$. \\
	\textbf{Output:} An adversarial raindrop example $x_{adv}$.\\
	
	\begin{algorithmic}[1]
		
		\For{$n=1\rightarrow N$}
		\State Sample an noise vector $z_{n}=\mathcal{N} \left( \textbf{0},\textbf{I} \right) $
		
		\For{$t=1\rightarrow T$}
		\State Generate the raindrop style sample
		 $x_{n}^{t}=G\left( x,z_{n}^{t} \right)$
		\State $x_{n}^{t} = clip(x_{n}^{t},0,1)$
		\If{$f\left( x_{n}^{t} \right) \ne y$ }
		\State $x_{adv}=x_{n}^{t}$
		\State \textbf{return}  $x_{adv}$
		\EndIf
		\State Calculate the gradient $g_{t}=\nabla _{z_{t}} L_{TC}\left(x_{n}^{t} ,y \right) $
		\State Update $z_{n}^{t+1}$ by applying the sign of gradient
		\begin{equation}
			z_{n}^{t+1}=z_{n}^{t}+\alpha _z\cdot sign\left( g_t \right)
		\end{equation}
		
		\EndFor
		\EndFor
		
	\end{algorithmic}
\end{algorithm} 

\section{Experiments}
\label{experiment}
To demonstrate the effectiveness of our physically and digitally generated raindrops in adversarial attack against DNN models, we conduct three groups of experiments. The first group of experiments in Sec.~\ref{realness} are to show that the raindrop images crafted by our AdvRD scheme not only look like the realistic ones, but also distribute closely to the realistic raindrops in terms of a metric of statistical analysis. The second group of experiments in Sec.~\ref{digital-attack} are carried out to compare the performances of our adversarial raindrops and some conventional adversarial attacking methods. In the third group of experiments in Sec.~\ref{physical-attack}, we show that adversarial training with the raindrop images generated by AdvRD can improve the robustness of DNN models to real-world raindrop attack.  

\subsection{Setup}
\label{setup}

\noindent \textbf{Dataset.} We train our quasi-GAN architecture on the Raindrop Removal (RDR)~\cite{raidrop_removal_cvpr2018} dataset, which contains 1119 pairs of images, with various real-world background scenes and raindrops. All other experiments are carried out on three datasets, NIPS-17~\cite{NIPS-Competition}, and two traffic sign recognition datasets, Tsinghua-Tencent 100K (TT-100K)~\cite{TT-100K_traffic_sign_cvpr2016} and GTSRB~\cite{GTSRB}. The NIPS-17 dataset was released in the NIPS 2017 competition on Defenses against Adversarial Attacks, which contains 1000 labeled images with a resolution of $299\times299\times3$. It has been wildly used in many previous works~\cite{digital_transfer_3d22D,digital_transfer_nips2021}. TT-100K contains 45 different Chinese road sign classes, and GTSRB has 43 different German road sign classes.

\noindent \textbf{Implement Details.} The quasi-GAN architecture is trained using the Adam optimization algorithm~\cite{Adam}. The transfer classifier in Fig.~\ref{fig:total} is a Resnet50, which is pre-trained on the Imagenet~\cite{ImageNet}. We set the parameter $\eta$ in Eq.~(\ref{eq:untarget Lben}) as $\eta=2$. The prior hyper-parameter $\alpha _1\sim\alpha _3$ are set to 100, 100, and 0.8, respectively.
The dimension $d$ of latent variable $z$ is set to be 64. In our AdvRD scheme, the noise sampling number is set to 25, the iteration number T is 10, and the step size $\alpha _z$ is 0.05. 
\subsection{Authenticity Evaluation}
\label{realness}

\begin{figure}[!t]
	\centering
	\setlength{\belowcaptionskip}{-0.4cm}
	\includegraphics[width=\linewidth]{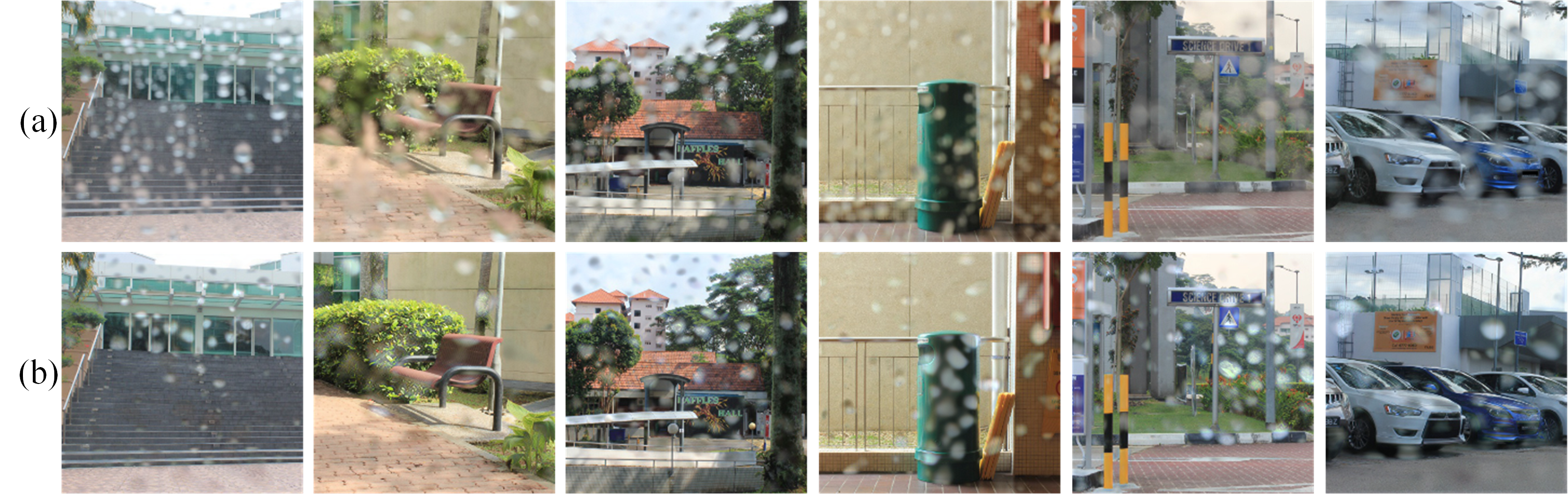}
	\caption{Visual similarity of (a) real-world raindrop images, and (b) adversarial raindrop images generated by our AdvRD scheme.}	
	\label{fig:reality}
\end{figure}

Using the clean images in the RDR dataset, we generate adversarial raindrop images, which look similar to their corresponding real-world raindrop images. Fig.~\ref{fig:reality} shows the visual similarity between real-world raindrops and those generated by our AdvRD scheme. Moreover, we employ a statistical metric, Fr$\acute{\text{e}}$chet Inception Distance (FID)~\cite{FID_distance}, to measure the distribution similarity between our adversarial raindrops and real-world raindrops.  

Basically, the FID metric was proposed to measure the difference of two Gaussian distributions $g1$ and $g2$, whose mean and covariance matrix are supposed to be ($m_1$,$C_1$) and ($m_2$,$C_2$), respectively. Then, the FID of $g1$ and $g2$ is defined as:
\begin{scriptsize}
\begin{equation}	
	FID(g1,g2)=\left\| m_1-m_2 \right\| _{2}^{2}+Tr\left( C_1+C_2-2\left( C_1C_2 \right) ^{1/2} \right). 
	\label{eq:FID}		
\end{equation}
\vspace{-12pt}
\end{scriptsize}

To estimate the FID value of the adversarial raindrops and realistic raindrops, we randomly divide the RDR dataset into two disjointed subsets, $r1$ and $r2$, with roughly same size, and then, use our AdvRD algorithm to generate a set of adversarial raindrop images, denoted as $f1$, from the clean images in $r1$. The values of FID($P_{r1}$,$P_{r2}$) and FID($P_{f1}$,$P_{r2}$) are calculated for each random sampling. We repeat the experiments 10 times, and Tab.~\ref{tab:RFID} presents the results, in which RFID is the ratio of FID($P_{f1}$,$P_{r2}$) and FID($P_{r1}$,$P_{r2}$). It can be seen that the values of RFID are very close to 1, indicating that the distribution difference between adversarial raindrops and realistic raindrops is almost the same as that between realistic raindrops. So, we can say that the raindrops generated by AdvRD almost have the same distribution as the realistic raindrops.
\begin{table}[!t]\small
	\centering
	\caption{Performance of AdvRD on the reality.}
	\begin{tabular}{cccc}
		\toprule
		Metrics &FID$\left( P_{f1},P_{r2} \right) $  & FID$\left( P_{r1},P_{r2} \right) $  & RFID \\
		\midrule
		 value    & 34.03$\pm$0.55 & 33.28$\pm$0.50 & 1.023$\pm$0.012 \\
		\bottomrule
	\end{tabular}
	
	\label{tab:RFID}
\end{table}        
\subsection{Attacking with Adversarial Raindrops}
\label{digital-attack}

We conduct the second group of experiments to evaluate the performance of the adversarial raindrops crafted by our AdvRD in attacking some typical DNN models. To do so, we compare our AdvRD method with six popular adversarial attacking approaches, namely, FGSM~\cite{FGSM}, BIM~\cite{BIM}, MIM~\cite{MI-FGSM}, DIM~\cite{DIM}, TIM~\cite{TIM} and NIM~\cite{NIM}, in terms of Attacking Success Rate (ASR) on the NIPS-17 dataset. The perturbation budget $\epsilon$ for these attacking approaches is set to be 16/255. 

In the case of white-box scenario, the target DNN models are chosen to be four standardly trained models: Inception-v3 (Inc-v3)~\cite{InceptionV3}, Inception-v4 (Inc-v4), Inception-Resnet-v2 (IncRes-v2)\cite{Inceptionv4}, and Resnet-v2-101 (Res-101)~\cite{Resnet}, and two robust models: Rob-ResNet50 (Rob-Res50)~\cite{PGD}, and ens-adv-Inception-ResNet-v2 (IncRes-v2$\rm _{\text{ens}}$)~\cite{Ens_incepton}. For black-box attacking, we choose three standardly trained networks, Inception-v4, Inception-Resnet-v2~\cite{Inceptionv4}, and Resnet-v2-101~\cite{Resnet}, and three adversarially trained models, ens3-adv-Inception-v3 (Inc-v3$\rm _{\text{ens3}}$), ens4-adv-Inception-v3 (Inc-v3$\rm _{\text{ens4}}$), and ens-adv-Inception-ResNet-v2~\cite{Ens_incepton}. For all the black-box attacks, except AdvRD, adversarial examples are generated and transferred from Inception-v3.

%

Experimental results are presented in Tab.~\ref{tab:digital-ASR-w} and Tab.~\ref{tab:digital-ASR-b}, for white-box attacks and black-box attacks, respectively. We see that 1) In the white-box scenario, The attacking ability of AdvRD raindrops is weaker in terms of ASR than that of the adversarial examples crafted by the gradient-based methods, except FGSM; 2) In black-box scenario, AdvRD outperforms other gradient-based methods by a big margin of ASR in most cases. Even for the three robust target models, which are pre-trained by adversarial training, our AdvRD raindrops still achieve more than 50\% ASR, remarkably higher than the gradient-based methods do. This may imply that the conventional gradient-based adversarial training does not work in defense against adversarial raindrops.     

\begin{table*}[!t]
	\caption{The white-box attack success rates (\%) $\uparrow$ on four undefended models and two adversarially trained models by various attacks.} 
	\centering
	\begin{tabular}{cccccccc}
		\toprule
		Methods & Inc-v3 & Inc-v4 & IncRes-v2 &Res-101   & Rob-Res50 & IncRes-v2$\rm _{\text{ens}}$   & Average \\
		\midrule
		FGSM  & 80.0  & 85.1 & 60.1 & 87.8  & 86.1 & 30.7 & 71.6 \\
		BIM   & \textbf{100.0} & \textbf{99.4} & \textbf{99.7} & \textbf{100.0} & \textbf{92.4} & \textbf{97.5} & \textbf{98.2} \\
		MIM   & \textbf{100.0} & 99.3 & 98.7 & 99.9  & 91.5 & 97.2 & 97.8 \\
		DIM   & 99.7  & \textbf{99.4} & 96.9 & 99.9  & 89.6 & 89.6 & 95.9 \\
		TIM   & 99.5  & 99.2 & 97.3 & 99.7  & 88.4 & 92.4 & 96.1 \\
		NIM   & \textbf{100.0} & \textbf{99.4} & 99.0 & 99.9  & 91.7 & 97.3 & 97.9 \\
		AdvRD & 84.2  & 91.0 & 89.7 & 88.1  & 88.1 & 89.5  & 88.4 \\
		\bottomrule
	\end{tabular}
	\label{tab:digital-ASR-w}
\end{table*}

\begin{table*}[!t]
	\caption{The black-box attack success rates (\%) $\uparrow$ on three undefended models and three adversarially trained models by various attacks.}
	\centering
	\begin{tabular}{cccccccc}
		\toprule
		Methods & Inc-v4 & IncRes-v2 & Res-101 & Inc-v3$\rm _{\text{ens3}}$ & Inc-v3$\rm _{\text{ens4}}$ & IncRes-v2$\rm _{\text{ens}}$ & Average \\
		\midrule
		FGSM    & 31.9   & 29.6      & 30.9    & 17.3       & 14.8       & 9.4          & 22.3    \\
		BIM     & 25.4   & 17.7      & 19.0    & 12.7       & 13.5       & 7.5          & 16.0    \\
		MIM     & 46.3   & 43.3      & 38.2    & 20.0       & 18.0       & 11.7         & 29.6    \\
		DIM     & 38.5   & 30.6      & 27.2    & 15.3       & 15.2       & 8.5          & 22.6    \\
		TIM     & 50.1   & 43.2      & 39.4    & 29.2       & 27.3       & 18.8         & 34.7    \\
		NIM     & 52.9   & \textbf{51.3}      & 41.4    & 20.3       & 18.1       & 11.1         & 32.5    \\
		AdvRD   & \textbf{53.9}   & 47.6      & \textbf{58.8}    & \textbf{62.4}       & \textbf{63.4}       & \textbf{52.4}         & \textbf{56.4} \\
		\bottomrule  
	\end{tabular}
	\label{tab:digital-ASR-b}
\end{table*}

\begin{figure*}[!t]
	\centering
	
	\includegraphics[width=0.95\linewidth]{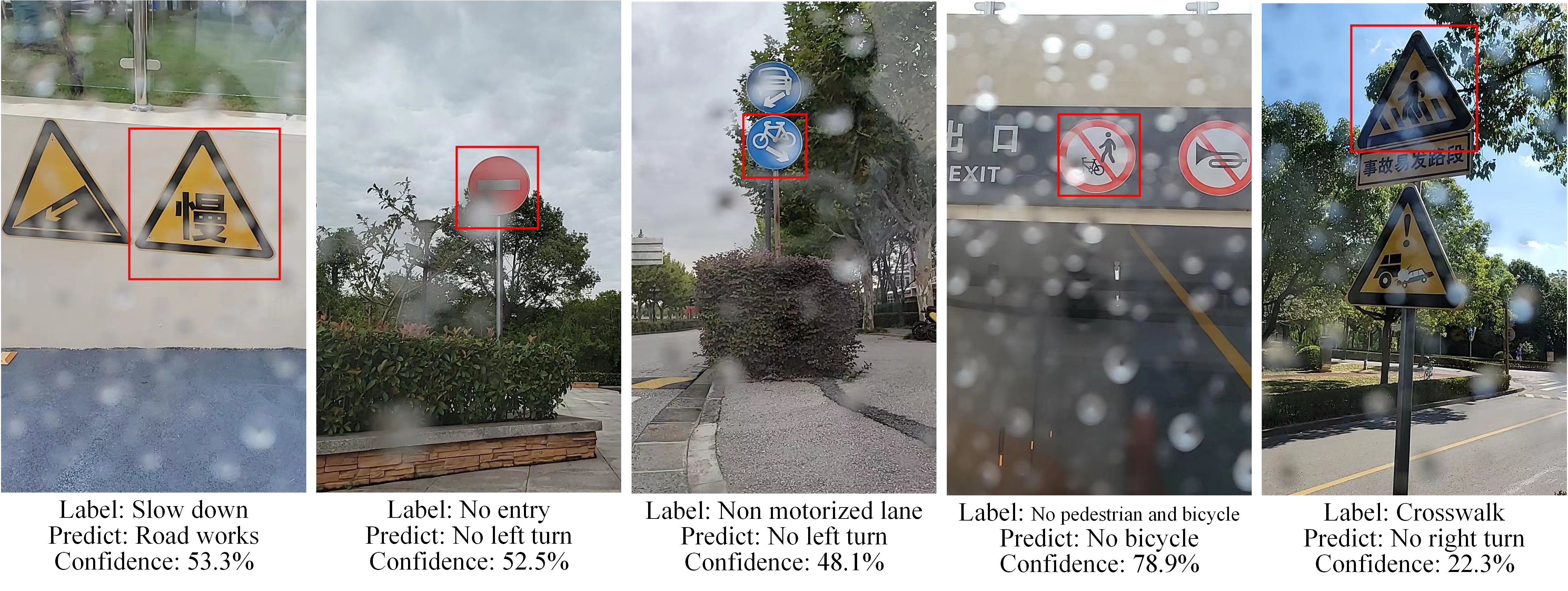}
	\setlength{\abovecaptionskip}{0.1cm}
	\setlength{\belowcaptionskip}{-0.4cm}
	\caption{Adversarial raindrop examples captured in the real-world and the corresponding classification results.}
	\label{fig:phybig}
\end{figure*}

\begin{table}
	\centering
	\caption{Attack success rate (\%) of real-world raindrop on three datasets.}
	\begin{tabular}{cccc}
		\toprule
		Dataset & NIPS-17 & TT-100K & GTSRB \\
		\midrule
		ASR     & 92.0     & 54.0    &59.0\\
		\bottomrule
	\end{tabular}
	\vspace{-9pt}
	\label{tab:physical-ASR}
\end{table}

\subsection{Defense Against Adversarial Raindrops}
\label{physical-attack}
In this part, we conduct experiments to show that real-world raindrops could become adversarial raindrops to DNN models, and then, we provide a defense method against adversarial raindrops.     

\noindent \textbf{Find real-world adversarial raindrops.}~In Sec.~\ref{real-world raindrop}, we describe how to find the real-world adversarial raindrop images to a DNN model. Given an image, the adversarial raindrop image, if found in 5 second video, is physically crafted adversarial example. Fig.~\ref{fig:phybig} shows five real-world adversarial raindrop images that mislead a DNN model pre-trained on TT-100K. We conduct experiments on three datasets, NIPS-17, TT-100K, and GTSRB, to estimate the probability that we can successfully find the adversarial raindrop image. We view this probability as the Attack Success Rate (ASR) of the real-world adversarial raindrops in attacking a DNN model. 

Tab.~\ref{tab:physical-ASR} lists the estimated probability or ASR that our real-world raindrops mislead the DNN models. We see that the values of ASR on the datasets are relatively high (over 50\%), indicating that DNN models are vulnerable not only to digital perturbations of adversarial examples, but also to the natural perturbations of raindrops. Especially, on NIPS-17, the ASR value reaches 92\%, which means that almost all the images in the dataset can naturally become an adversarial example, only if a few of raindrops are sprayed on them.    

\noindent \textbf{Defense against Adversarial Raindrops.} Adversarial training is usually considered as the most effective way to increase robustness of DNN models to adversarial perturbations. To defend against adversarial raindrops, we investigate the effectiveness of applying adversarial training to improve DNNs' robustness. Since adversarial training needs large number of adversarial examples, and it is difficult to obtain enough real-world adversarial raindrop images, we use our AdvRD raindrops instead in the experiments of adversarial training. Specifically, in each epoch of our adversarial training, we randomly select half of the training data to generate AdvRD raindrop images, and combine them with the other half of clean data to train the model. Tab.~\ref{tab:AT} gives the experimental results,  upper half of which is for standard training, and the lower half is for adversarial training. It can be seen that adversarial training with our AdvRD raindrops significantly reduce the ASR values for both digital and physical raindrop attacking. Like the conventional adversarial training, our adversarial training in the experiments also decreases the recognition accuracy on clean images of NIPS-17. But, surprisingly, it even improves the recognition accuracy on clean samples of datasets TT-100K and GTSRB.  
\begin{table}
	\centering
	\caption{Performance comparison of models with and without adversarial training.}
	\begin{tabular}{lccc}
		\toprule
		Model             & Acc. & Dig. ASR & Phy. ASR \\
		\midrule
		NIPS-Resnet50     & 76.13    & 64.5        & 92.0         \\
		TT-Resnet18       & 98.11    & 72.6        & 54.0         \\
		GTSRB-Resnet18    & 97.52    & 66.6        & 59.0            \\
		\midrule
		NIPS-Resnet50$_{\rm rob}$  & 73.37    & 29.0        & 69.0         \\
		TT-Resnet18$_{\rm rob}$    & 99.67    & 18.2        & 27.0        \\
		GTSRB-Resnet18$_{\rm rob}$ & 98.99    & 23.5        & 37.0            \\
		\bottomrule
	\end{tabular}
	
	\vspace{-7pt}
	\label{tab:AT}
\end{table}    
%

\subsection{Why Adversarial Raindrops Work}
\label{discussion}
\begin{figure}[!t]
	\centering
	
	\includegraphics[width=\linewidth]{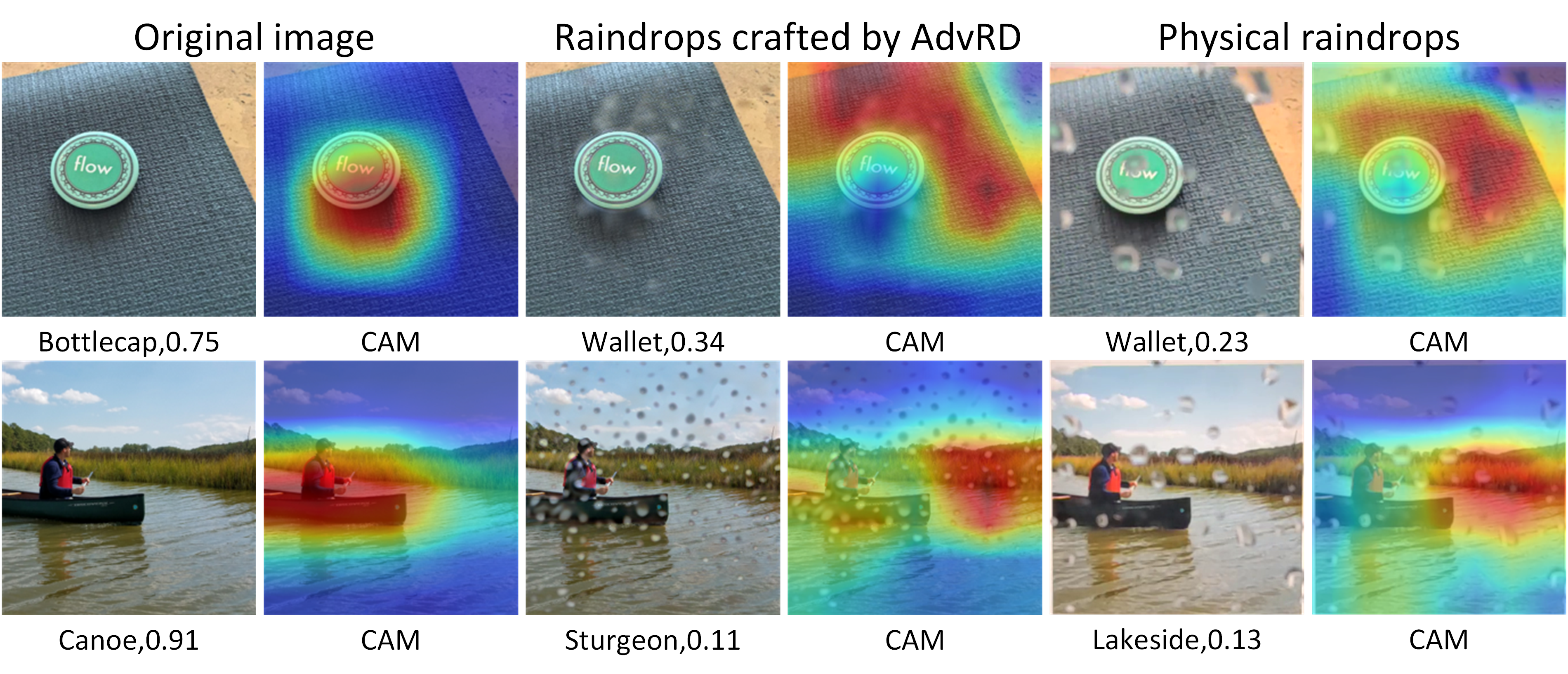}
	\setlength{\abovecaptionskip}{-0.2cm}
	\setlength{\belowcaptionskip}{-0.6cm}
	\caption{CAM for images.}
	\label{fig:CAM}
\end{figure}

To understand the mechanism that raindrops can mislead DNN models,, we visualize the CAM attention~\cite{CAM}, before and after adding the adversarial raindrops. Fig.~\ref{fig:CAM} shows the CAM attention of a DNN model on clean images, AdvRD crafted raindrop images, and physical adversarial raindrop images, respectively. We see that, though raindrops only perturb sparse pixels, the attention maps are substantially disturbed. It is worth noting that the objects in the raindrop perturbed images are visually complete and clear, which implies that image degeneration may not be the main reason for misclassification. Raindrops are responsible for disturbing the CAM attention maps. More real-world adversarial raindrops and their CAM are included in supplementary material.

\subsection{Ablution Study}
\label{ablution}

\begin{figure}[!t]
	\centering
	
	\includegraphics[width=0.9\linewidth]{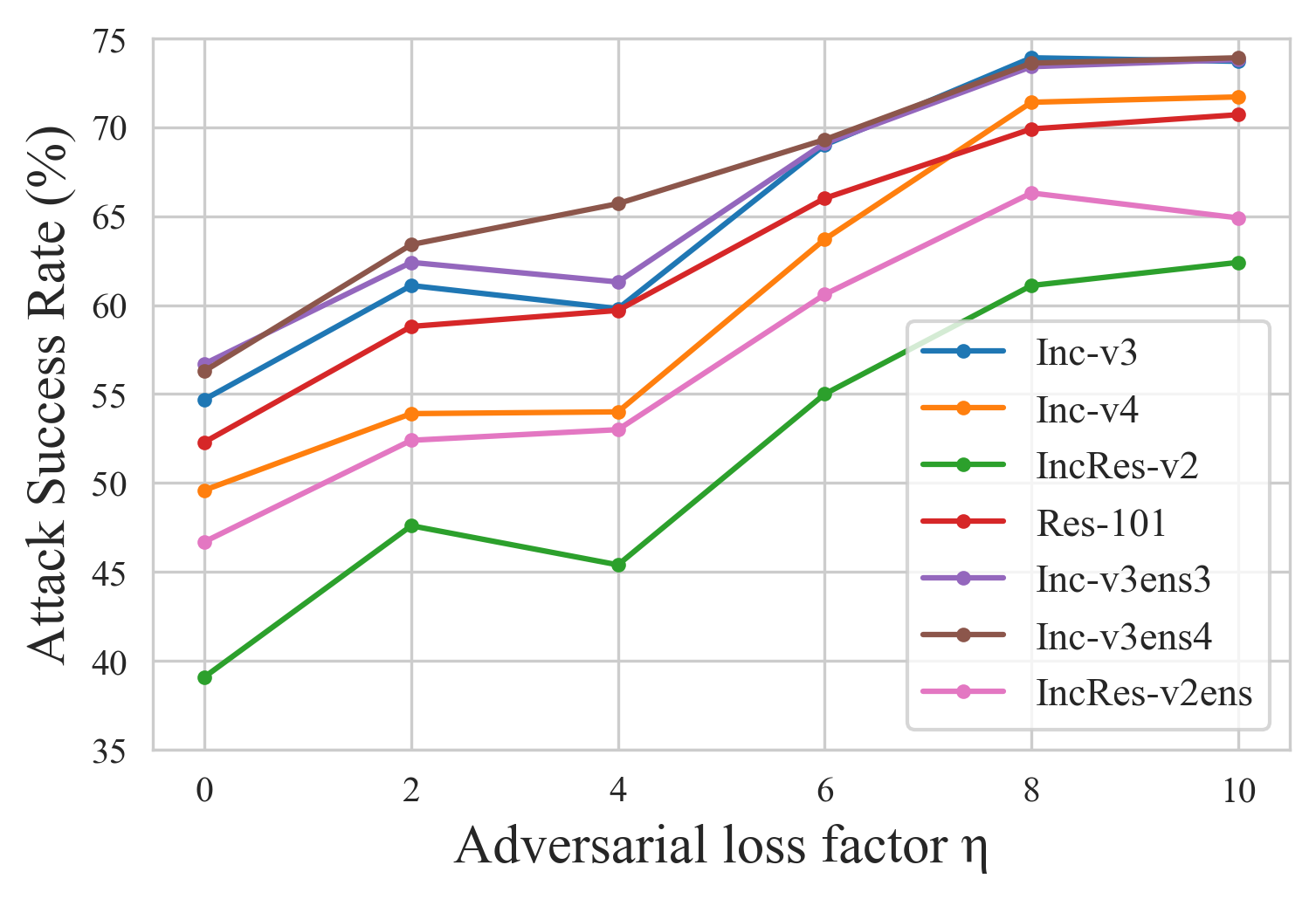}
	\setlength{\abovecaptionskip}{0.0cm}
	\setlength{\belowcaptionskip}{-0.3cm}
	\caption{The ASR curves of AdvRD trained with different $\eta$.}
	\label{fig:EOY}
\end{figure}

\noindent \textbf{The parameter $\eta$.} The main effect of $\eta$ in Eq.~(\ref{eq:untarget Lben}) is to balance the trade-off between adversarial strength and the reality of synthetic raindrops. A generator trained with a larger $\eta$ tends to fool the target classifier more easily. However, a too large $\eta$ may make the generator focus on cheating the classifier rather than generate realistic raindrop images. We test the effect of $\eta$ by setting it from 2 to 10 with a step size of 2. The ASR curves for seven target models are shown in Fig.~\ref{fig:EOY}. Tab.~\ref{tab:EOY} presents the RFID values corresponding to different $\eta$. As can be seen in Fig.~\ref{fig:EOY}, a generator fine-tuned with larger $\eta$ achieves higher values of ASR, which indicates that increasing $\eta$ can improve the attack capability of AdvRD. On the other hand, we see from Tab.~\ref{tab:EOY} that a larger $\eta$ causes a higher value of RFID, which means the sacrifice of reality.  

\begin{table}
	\centering
	\caption{Reality of proposed AdvRD trained with different $\eta$.}
\begin{tabular}{cccccc}
	\toprule
	$\eta$& 2     & 4     & 6     & 8     & 10    \\
	\midrule
			
	RFID   & 1.023 & 1.045 & 1.046 & 1.065 & 1.180 \\
	\bottomrule
\end{tabular}
	\vspace{-5pt}
	\label{tab:EOY}
\end{table}

\noindent \textbf{Noise sampling number $N$.} Obviously, increasing $N$ will improve the attack strength but impair the attack efficiency, since the attacker queries the target model more times to search an adversarial example. We set $N$ from 15 to 35 with a step size of 5 to test the influence of noise sampling number. The values of ASR and running time to finish attacking  NIPS-17 are shown in Tab.~\ref{tab:EON}. We can observe that both the attack strength and running time are positively related to $N$. So in practice, we set $N=25$ to balance the attack strength and efficiency.   

\begin{table}
	\centering
	\caption{Ablation study of noise sample number $N$.}
	\begin{tabular}{cccccc}
		\toprule
		N             & 15   & 20   & 25   & 30   & 35   \\
		\midrule
		ASR (\%)      & 59.7 & 61.2 & 63.4 & 65.2 & 67.1 \\
		Running Time (s)  & 445  & 535  & 636  & 722  & 792  \\
		
		\bottomrule
	\end{tabular}
	\vspace{-10pt}
	\label{tab:EON}
\end{table}


\section{Conclusion}
In this paper, we study the adversarial examples caused by natural raindrops, and present a new approach to generate adversarial raindrops in the digital domain, using a quasi-GAN technique. The generated raindrop images are very similar to the real-world raindrop images, from viewpoints of human vision and statistical analysis. More importantly, they perform strong adversarial attack to the state-of-the-art DNNs. We also show that the adversarial training using our AdvRD images can significantly improve the robustness of DNNs to the real-world raindrop attacks.



{\small
\bibliographystyle{ieee_fullname}
\bibliography{egbib}
}

\end{document}